\documentclass[lettersize,journal]{IEEEtran}

\usepackage{hyperref}       
\hypersetup{colorlinks=true,urlcolor=black}
\usepackage{url}            
\usepackage{booktabs}       
\usepackage{amsfonts}       
\usepackage{nicefrac}       
\usepackage{microtype}      
\usepackage{algorithm}
\usepackage{algorithmic}
\usepackage{graphicx}
\usepackage{epsfig}
\usepackage{amsmath}
\usepackage{amssymb}
\usepackage{amsthm}
\usepackage{subfigure}
\usepackage{booktabs} 
\usepackage{multirow}
\usepackage{etoolbox}
\usepackage{bbm}
\usepackage{sail}
\usepackage{color}
\usepackage{footnote}
\usepackage{enumitem}
\usepackage{times}
\usepackage{ragged2e}
\usepackage{xspace}
\usepackage{threeparttable}
\usepackage[export]{adjustbox}
\usepackage[table]{xcolor}
\usepackage{arydshln}

\def\shortname{DRVR\xspace}

\begin{document}
\title{Semantic Occupancy Prediction with Dual Range-Voxel Representation}
\author{Sitao Chen, Zhuangwei Zhuang, Hui Luo, Lizhao Liu, Qingyao Wu, Mingkui Tan, \IEEEmembership{Senior Member, IEEE}

\thanks{Sitao Chen, Zhuangwei Zhuang, Lizhao Liu, Qingyao Wu and Mingkui Tan are with the School of Software Engineering, South China University of Technology, Guangzhou 510006, China (e-mail: chensitao27@gmail.com; \{z.zhuangwei, selizhaoliu\}@mail.scut.edu.cn; \{qyw, mingkuitan\}@scut.edu.cn).}
\thanks{Hui Luo is with the State Key Laboratory of Optical Field Manipulation Science and Technology, Institute of Optics and Electronics, CAS, Chengdu 610209, China (e-mail: luohui19@mails.ucas.ac.cn).}
\thanks{Corresponding authors: Hui Luo; Mingkui Tan.}
}

\maketitle

\begin{abstract}
LiDAR-based 3D semantic occupancy prediction, which aims to provide accurate and comprehensive scene representation, is crucial for autonomous driving systems. As point clouds suffer from sparsity and incompleteness, leading to insufficient semantic learning and difficult occupancy perception, existing methods often stack multi-sweep point clouds to obtain dense spatial information. However, such a naive strategy also results in efficiency (\textit{e.g.}, additional computational burden) and robustness (\textit{e.g.}, pose transformation noise) concerns, which hinder their practical applications. In this work, we propose a Dual Range-Voxel Representation (\shortname) that leverages the range-view context and voxel-view geometry of single-sweep point clouds for 3D semantic occupancy prediction, eliminating the concerns associated with the multi-sweeps. Specifically, we use the range-view encoder to extract the compact context of the scene. To fully exploit the spatial information, we design a geometry-aware voxel-view encoder that extracts multi-scale voxel-view features separately and combines them for better geometric occupancy prediction. Moreover, we propose a range-voxel fusion module to cooperate range- and voxel-view features via voxel-to-range and range-to-voxel fusions. Extensive experiments on nuScenes-Occupancy, SemanticKITTI and SemanticPOSS show the superiority of our method. Especially on nuScenes-Occupancy, our single-sweep \shortname achieves 5.4\% improvement in mIoU and 2.1$\times$ acceleration compared to the multi-sweep method. Our source code is available at \url{https://github.com/chenst27/DRVR}.
\end{abstract}


\section{Introduction}
\label{Introduction}
\IEEEPARstart{A}{ccurate} and comprehensive 3D perception is critical in autonomous driving systems~\cite{geiger2012we, huang2018apolloscape, liao2022kitti, sun2020scalability}, providing detailed environmental information for subsequent path planning, decision-making, and motion control. One of the important tasks in 3D perception is 3D semantic occupancy prediction. Unlike 3D object detection~\cite{bai2022transfusion, chen2017multi, li2022deepfusion, yin2021center} and LiDAR segmentation~\cite{hong2021lidar, yan20222dpass, zhou2021panoptic, zhuang2021perception,tan2024epmf} that focus on foreground objects or discrete point clouds, 3D semantic occupancy prediction represents continuous 3D space through dense grid cells and estimates both the occupancy state and semantics of each voxel, achieving holistic scene understanding~\cite{li2023sscbench, tian2024occ3d, zhuang2024robust, ouyang2024linkocc, lu2024lidar, yang2024adaptiveocc, wang2025occ, yang2025daocc, xiao2024semantic}.

\begin{figure}[t]
    \centering
    \includegraphics[width=1.0\linewidth]{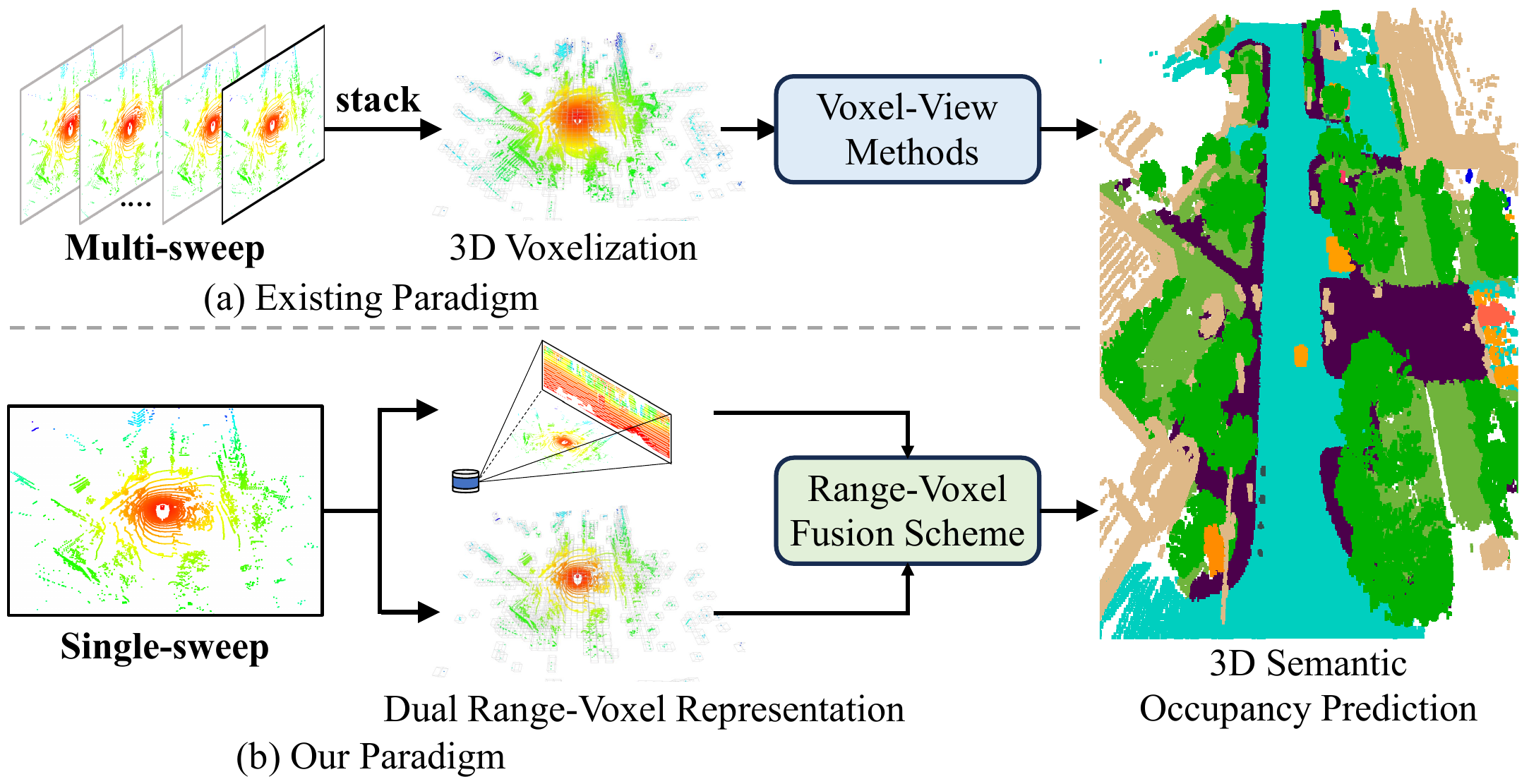}
    \caption{Comparisons of our paradigm to existing paradigm~\cite{wang2023openoccupancy, wang2024occgen}. The existing paradigm stacks and voxelizes multi-sweep point clouds to obtain the geometric representation, which is fed into voxel-view methods for prediction. In contrast, our paradigm generates the dual range-voxel representation from single-sweep point clouds via spherical projection and 3D voxelization and applies the range-voxel fusion scheme for semantic occupancy prediction.}
    \label{fig:paradigm_cmp}
\end{figure}

Existing semantic occupancy prediction methods can be mainly categorized into camera-based methods~\cite{cao2022monoscene, huang2023tri, li2023voxformer, zhang2023occformer} and LiDAR-based methods~\cite{roldao2020lmscnet, wang2023openoccupancy, xia2023scpnet, yan2021sparse}. Camera-based methods leverage the images with rich appearance information (\eg, RGB color and texture) to predict 3D occupancy. However, they are difficult to make accurate geometric predictions due to the lack of depth sensing~\cite{pan2024co, zhang2024radocc}. In contrast, LiDAR-based methods show advantages in geometric modeling using reliable and accurate spatial information from point clouds. However, point clouds suffer from sparsity and incompleteness, where the former results in insufficient semantic representation learning and the latter poses challenges for occupancy perception. To infer the coherent semantics and geometry of 3D space, existing methods~\cite{pan2024co, wang2023openoccupancy, wang2024occgen, zhang2024radocc} typically stack multi-sweep point clouds (\ie, current and past moments) to obtain dense information (See Figure~\ref{fig:paradigm_cmp}(a)). These methods struggle to meet the efficiency and robustness requirements of real-life applications, where limited computational resources and noisy pose transformation\footnote{See Sec. \ref{robustness_to_noise} for more details.} may lead to sub-optimal and unsatisfactory performance.

In this work, we aim to predict continuous semantic occupancy from single-sweep sparse and incomplete point clouds, eliminating the efficiency and robustness concerns raised by the dependency on the multi-sweeps. Motivated by existing studies on range-view LiDAR semantic segmentation~\cite{cortinhal2020salsanext, milioto2019rangenet++}, we notice that using spherical projection on point clouds can obtain the spatially continuous and contextually compact range-view representation, which facilitates the model to capture the semantic features (See Figure~\ref{fig:paradigm_cmp}(b)). However, spherical projection inevitably destroys the geometric structure~\cite{liu2023uniseg, xu2021rpvnet}. In contrast, while regular voxel-view representation preserves spatial information and enhances geometric perception, it struggles to capture semantic information. We believe that integrating these two views can enable the model to capture both semantics and geometry within sparse point clouds.

Based on this intuition, we propose a Dual Range-Voxel Representation (\textbf{\shortname}) that leverages the range-view context and voxel-view geometry to enhance semantic understanding and geometric perception for 3D semantic occupancy prediction. Specifically, we represent point clouds as a range-view image and use the range-view encoder to extract the 2D feature, which captures the compact contextual information of the scene. To fully exploit the spatial information of point clouds, we represent them as multi-scale voxels and further propose the geometry-aware voxel-view encoder to predict geometric occupancy and generate voxel-view features for the occupied voxels. For range- and voxel-view fusion, we propose the collaborative range-voxel fusion module consisting of voxel-to-range and range-to-voxel fusions. Within the voxel-to-range fusion, we apply the occupied voxels as queries to dynamically aggregate the range-view feature of interest via deformable cross-attention mechanism~\cite{zhu2020deformable}, benefiting from its rich context. In the range-to-voxel fusion, we transfer range-view features into point-view, thus leveraging the spatial information of point clouds. Then, the point-view features are voxelized and added to the voxel-view features of occupied voxels for fusion, which are further fed into 3D sparse convolutions for enhancement. Lastly, we merge the dual-path fused features for semantic occupancy prediction. The main contributions of this work are summarized as follows.
\begin{itemize}
  \item We propose a Dual Range-Voxel Representation (\textbf{\shortname}) that leverages range-view context and voxel-view geometry to achieve 3D semantic occupancy prediction from single-sweep sparse point clouds. Our method effectively overcomes the efficiency and robustness limitations inherent in methods that depend on multi-sweep point clouds.
  \item We design a geometry-aware voxel-view encoder that fully exploits spatial information from sparse point clouds through multi-scale feature extraction and BEV fusion, significantly enhancing geometric awareness and thereby enabling more accurate occupancy prediction.
  \item We introduce a collaborative range-voxel fusion module that effectively cooperates range- and voxel-view features via voxel-to-range and range-to-voxel fusions. Extensive experimental results on three benchmarks consistently demonstrate the superior performance of our method. 
\end{itemize}

\section{Related work}
\label{Related_work}
\subsection{Camera-based occupancy prediction.}
Camera-based occupancy prediction takes single- or multi-view images to reason the geometry and semantics of each voxel in the 3D scene, which is dominated by two mainstream paradigms. One paradigm exploits forward projection~\cite{philion2020lift} that projects image features into Bird’s Eye View (BEV) or voxel space based on the predicted pixel-wise depth distribution~\cite{huang2021bevdet, liu2023bevfusion, tang2024sparseocc, zhang2024radocc, zhang2023occformer}. Another paradigm follows backward projection~\cite{wang2022detr3d} that applies learnable queries to aggregate image features through 2D-to-3D spatial mapping relationship~\cite{jiang2024symphonize, li2022bevformer, li2025viewformer, liu2024fully, lu2023octreeocc, wang2024panoocc, wang2024h2gformer, wei2023surroundocc}. MonoScene~\cite{cao2022monoscene} is a pioneering work that infers semantic scene representation from a single image via 2D and 3D UNets. OccFormer~\cite{zhang2023occformer} presents a dual-path transformer to enhance voxel-view feature extraction. TPVFormer~\cite{huang2023tri} proposes a tri-perspective view (TPV) representation to describe the 3D surroundings. VoxFormer~\cite{li2023voxformer} uses an MAE-like architecture~\cite{he2022masked} to generate dense voxels from sparse ones. Symphonies~\cite{jiang2024symphonize} further introduces instance queries to exploit instance semantics and scene context. CGFormer~\cite{yu2024context} proposes a context and geometry aware voxel transformer for view transformation. Camera-based methods can effectively transfer 2D image semantics to 3D space. However, they underperform LiDAR-based methods in occupancy prediction due to ambiguous depth sensing.

\subsection{LiDAR-based occupancy prediction.}
LiDAR-based occupancy prediction aims to reconstruct the continuous 3D volumetric scene from sparse point clouds. LMSCNet~\cite{roldao2020lmscnet} completes the semantic scene through a 2D backbone and a 3D segmentation head. S3CNet~\cite{cheng2021s3cnet} applies different branches to predict the BEV semantic map and 3D scene representation and merges them via post-fusion. JS3C-Net~\cite{yan2021sparse} first performs point cloud segmentation and then feeds point-view segmentation into the SSC and point-voxel interaction modules for voxel-view feature generation. SSA-SC~\cite{yang2021semantic} combines 2D and 3D networks for 2D completion and 3D segmentation simultaneously and fuses them at multiple scales. SSC-RS~\cite{mei2023ssc} extracts multi-level semantic context and multi-scale geometric structures and fuses them via an adaptive representation fusion module. VPNet~\cite{wang2024voxel} introduces confident voxels to capture diverse possibilities of voxels and implicitly model the semantic uncertainty inherent in them. L-CONet~\cite{wang2023openoccupancy} feeds voxelized point clouds into the 3D encoder and decoder to extract voxel-view features for coarse-to-fine occupancy prediction. L-OccGen~\cite{wang2024occgen} follows the ``noise-to-occupancy" paradigm and uses point clouds voxel-view features as the condition for progressive refinement. However, considering the challenge of inferring the coherent semantics and geometry of 3D space from highly sparse point clouds, existing methods~\cite{wang2023openoccupancy, wang2024occgen, zhang2024radocc, zuo2023pointocc} tend to stack multi-sweep point clouds for dense spatial representations. In this work, we propose to leverage the range-view context and voxel-view geometry of single-sweep sparse point clouds to construct the semantic occupancy representation of the 3D scene.

\begin{figure*}[t]
    \centering
    \includegraphics[width=1.0\linewidth]{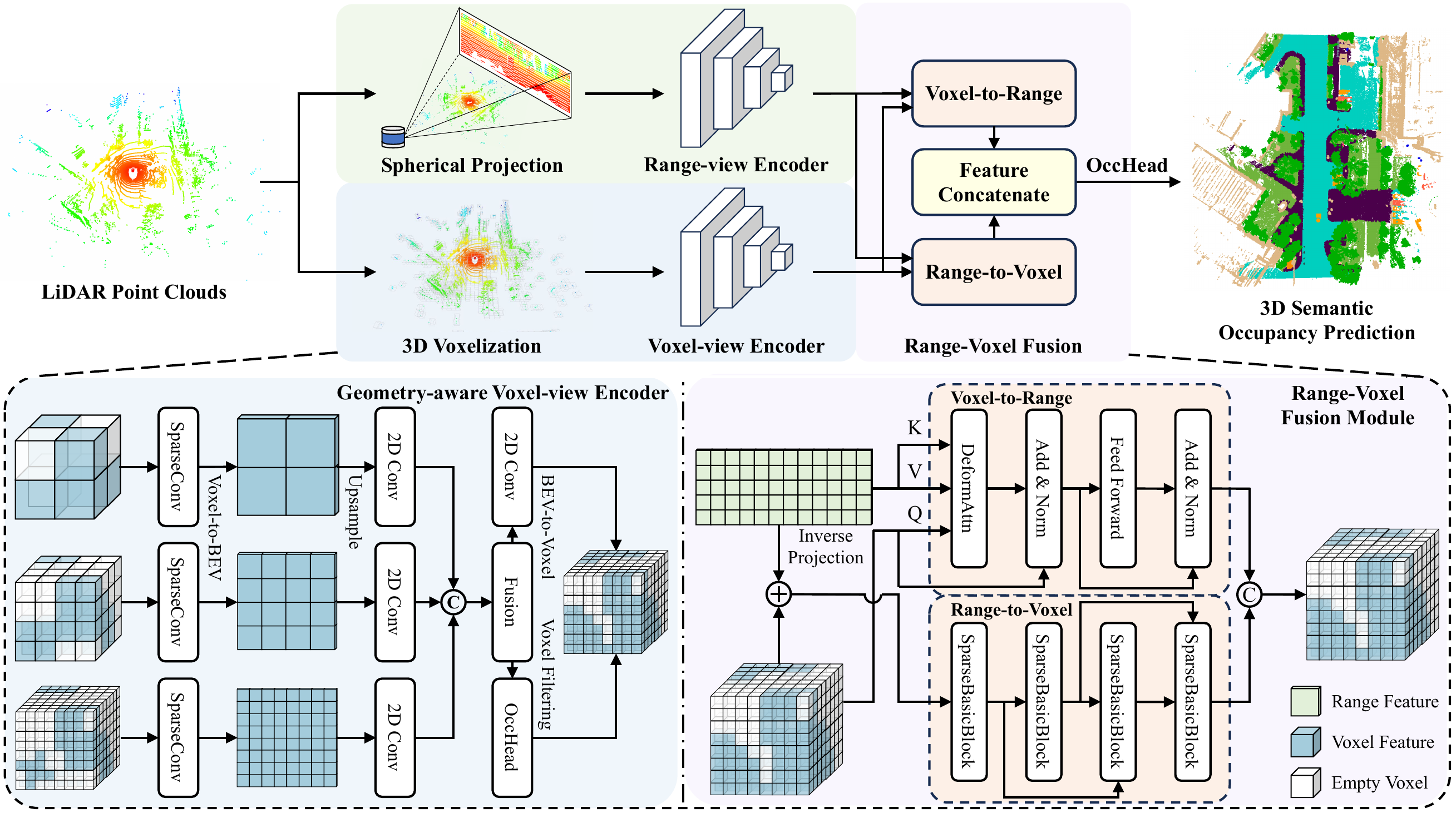}
    \caption{General architecture of \shortname.
    Firstly, we project the point clouds to obtain the range-view image via spherical projection and apply the range-view encoder to extract the context-aware feature. Meanwhile, we perform 3D voxelization on point clouds to obtain multi-scale voxel-view features and employ the voxel-view encoder to extract the geometry-aware feature. Secondly, the range-view and voxel-view features are fed into voxel-to-range and range-to-voxel fusions for feature integration. Lastly, geometry occupancy and semantic occupancy are supervised by occupancy labels.}
    \label{fig:arch_overview}
\end{figure*}

\section{Methodology}
\label{Method}
In this work, we propose a dual range-voxel representation (\textbf{\shortname}) that leverages the range-view context and voxel-view geometry to enhance semantic understanding and geometric perception for 3D semantic occupancy prediction. Specifically, as illustrated in Figure~\ref{fig:arch_overview}, our \shortname contains three components: context-aware range-view encoder (Section~\ref{sec:range_encoder}), geometry-aware voxel-view encoder (Section~\ref{sec:voxel_encoder}) and range-voxel fusion module (Section~\ref{sec:fusion_module}). Different from existing methods~\cite{wang2023openoccupancy, wang2024occgen} that stack multi-sweep point clouds for dense spatial information, we first perform spherical projection and 3D voxelization on single-sweep sparse point clouds to generate dual range-voxel representation. Then, we apply range- and voxel-view encoders for feature extraction. The range-view context and voxel-view geometry are efficiently fused via the range-voxel fusion module to predict semantic occupancy. The general scheme of our \shortname is depicted in Algorithm~\ref{alg:optimization}.

\subsection{Problem Definition}

Let $\mathbf{P} \in \mathbb R^{N_{point} \times 4}$ indicates a point cloud from LiDAR and $N_{point}$ denotes the number of points. Each point contains a 3D coordinate $(x, y, z)$ and an intensity value $i$. Given the input point cloud $\mathbf{P}$, the LiDAR-based semantic occupancy prediction task aims to predict the semantic class of each voxel in the voxel space $\mathbf{O} \in \mathbb R^{X \times Y \times Z}$ and indicate whether the voxel is empty or belongs to a specific semantic class $s\in\{0, 1, \dots, S\}$. Here, $S$ is the number of semantic classes, and $0$ often denotes the free voxel. $(X,Y,Z)$ denotes the resolution of 3D space.

\subsection{Context-aware Range-view Encoder}
\label{sec:range_encoder}

To mitigate the spatial discontinuity and contextual sparsity of single-sweep point clouds, we leverage the spatially continuous and contextually compact range-view representation for semantic representation learning. To obtain the input range-view image, we use spherical projection~\cite{cortinhal2020salsanext, milioto2019rangenet++} to project the LiDAR point cloud to the spherical coordinate. Specifically, each point $\mathbf{p} \in \mathbf{P}$ with a 3D coordinate $(x, y, z)$ can be mapped to a 2D coordinate $(u, v)$ on the range-view image by
\begin{equation}
\left(
\begin{matrix}
	u \\
	v
\end{matrix}
\right) =
\left(
\begin{matrix}
    \frac{1}{2}\left [ 1-\arctan\left ( y,x \right ) \pi^{-1} \right ]W_{img} \\
    \left [ 1 - \left (  \arcsin\left ( z, r^{-1} \right ) + |f_{down}| \right ) f_{v}^{-1} \right ]H_{img}
\end{matrix}
\right),
\label{eq:spherical_projection}
\end{equation}
where $H_{img}, W_{img}$ denote the height and width of the projected image, $r=\sqrt{x^2+y^2+z^2}$ indicates the range value of each point and $f_{v} = |f_{up}| + |f_{down}|$ refers to the vertical field-of-view of the LiDAR sensor. Similar to ~\cite{cheng2022cenet, cortinhal2020salsanext, milioto2019rangenet++}, we obtain a range-view image $\mathbf{I} \in \mathbb R^{5 \times H_{img} \times W_{img}}$ after spherical projection, where each pixel contains 5 channels $(x,y,z,i,r)$. If multiple points are projected onto the same pixel, only the feature of the point with the smallest range is preserved. The empty pixels are filled with zeros.

Given the input range-view image $\mathbf{I}$, we adopt the range-view encoder to extract the 2D feature, which captures the compact context information of the scene. The range-view feature $\mathbf{F}^{2D} \in \mathbb R^{C \times H_{f} \times W_{f}}$ is computed by 
\begin{equation}
    \label{eq:extract_range_feature}
    \mathbf{F}^{2D} = \text{RangeViewEncoder}(\mathbf{I}),
\end{equation}
where $C, H_{f}$ and $W_{f}$ denote the feature dimension, height and width of the feature map, respectively. 

\subsection{Geometry-aware Voxel-view Encoder}
\label{sec:voxel_encoder}
The geometric structure of point clouds is damaged by spherical projection~\cite{liu2023uniseg, xu2021rpvnet}. Hence, we apply 3D voxelization~\cite{zhou2018voxelnet, yan2018second} on raw point clouds to obtain regular voxel-view representation, preserving the spatial information. Specifically, given a point $\mathbf{p}$, its coordinates in the voxel grid can be calculated by
\begin{equation}
    \label{eq:voxelization}
    (i, j, k) = (\lfloor x/r_x \rfloor, \lfloor y/r_y \rfloor, \lfloor z/r_z \rfloor),
\end{equation}
where $r_x$, $r_y$, and $r_z$ are the voxel sizes along the X, Y, and Z axes. Then, the voxel-view features are obtained by aggregating point cloud features within the same voxel grid.

As shown in Figure~\ref{fig:arch_overview}, to fully exploit the geometric information of point clouds, we set different voxel resolutions to generate multi-scale voxel-view features $\mV = \{\bV_1, ..., \bV_l\}$, where $\bV_l \in \mmR^{N_{l} \times C}$ is the voxel-view feature in the $l$-th scale, followed by 3D sparse convolutions~\cite{spconv2022}. Considering the sparsity of voxel-view features, we propose to fuse multi-scale features at the BEV plane. Specifically, we first compress voxel-view features $\mV$ along the height dimension to obtain the BEV features $\mB = \{\bB_1, ..., \bB_l\}$, where $\bB_l \in \mmR^{C \times H_l \times W_l}$ is the BEV feature in the $l$-th scale. Then, we upsample the BEV features to the target resolution and apply 2D convolutions to fuse the multi-scale features. The fused BEV feature $\mathbf{F}^{BEV} \in \mmR^{C \times H \times W}$ is fed into the classifier to predict the occupancy $\hat{\mathbf{O}}_{BEV} \in \mmR^{(2 \times D) \times H \times W}$ and 2D convolutions for further feature extraction. We reshape the predicted occupancy $\hat{\mathbf{O}}_{BEV}$ to obtain voxel-view occupancy prediction $\hat{\mathbf{O}}_{geo} \in \mmR^{2 \times H \times W \times D}$ and split the BEV feature $\mathbf{F}^{BEV}$ to generate the voxel-view feature $\mathbf{F}^{Voxel} \in \mmR^{C^{\prime} \times H \times W \times D}$, where $C = C^{\prime} \times D$. Lastly, based on the predicted occupancy $\hat{\mathbf{O}}_{geo}$, we filter out the empty voxels and apply a lightweight MLP to generate the sparse voxel-view feature $\mathbf{F}^{3D} \in \mmR^{N_{voxel} \times C}$ of the occupied voxels. The simplified process is as follows:
\begin{equation}
    \label{eq:extract_voxel_feature}
    \mathbf{F}^{3D} = \text{VoxelViewEncoder}(\mV),
\end{equation}
where $N_{voxel}$ and $C$ denote the predicted number of occupied voxels and feature dimension, respectively.

\begin{algorithm}[t]
    \caption{General Scheme of \shortname}
    \begin{algorithmic}[1]
        \REQUIRE Training data $\{\bP, \bO\}$, \shortname model $M$.
        \WHILE{\textit{not convergent}}
            \STATE Project the point clouds $\bP$ to obtain the range-view representation $\bI$ by Eq.~(\ref{eq:spherical_projection}).
            \STATE Extract context-aware range-view feature $\bF^{2D}$ by Eq.~(\ref{eq:extract_range_feature}).
            \STATE Voxelize the point clouds $\bP$ to obtain the multi-scale voxel-view representation $\mV$ by Eq.~(\ref{eq:voxelization}).
            \STATE Extract geometry-aware voxel-view feature $\bF^{3D}$ by Eq.~(\ref{eq:extract_voxel_feature}) and compute geometric occupancy $\hat{\bO}_{geo}$.
            \STATE Obtain fused feature $\bF^{3D}_{fusion}$ by Eqs.~(\ref{eq:voxel2range}), (\ref{eq:range2voxel}) and (\ref{eq:fusion}) and compute semantic occupancy $\hat{\bO}_{sem}$.
            \STATE Compute the object function of geometric occupancy and semantic occupancy with labels $\bO$ by Eqs.~(\ref{eq:loss_geo}) and (\ref{eq:loss_sem}).
            \STATE Update model $M$ by minimizing the objective in Eq.~(\ref{eq:loss_overall}).
        \ENDWHILE
    \end{algorithmic}
    \label{alg:optimization}
\end{algorithm}

\subsection{Range-Voxel Fusion Module}
\label{sec:fusion_module}

After the multi-view feature encoder, we obtain the context-aware range-view and geometry-aware voxel-view features. For the range- and voxel-view features fusion, we propose the range-voxel fusion module that first performs voxel-to-range and range-to-voxel fusions and then integrates the fused features of different streams. 

\noindent\textbf{Voxel-to-range fusion.}
We project the real-world central coordinates of occupied voxels onto the range-view image via spherical projection, thus building the correspondence between them. A simple fusion method is to sample range-view features based on projected coordinates. However, it remains suboptimal due to the limited receptive field and rigid sampling. Inspired by camera-based methods~\cite{li2023voxformer, li2022bevformer}, we extend the deformable attention~\cite{zhu2020deformable} to our LiDAR-based method, which is capable of dynamically capturing features of interest. Specifically, given the feature map $F$, each query $q$ will aggregate the sampled features around the corresponding 2D reference point $p$ through 
\begin{equation}
    \label{eq:deformattn}
    \text{DeformAttn}(q, p, F) = \\
    \sum_{i=1}^{N_{h}} W_i \sum_{j=1}^{N_{s}} A_{ij} W_{i}^{'} F(p +  \triangle p_{ij}),
\end{equation}
where $N_{h}$ and $N_{s}$ denote the number of attention heads and sampled points, respectively. $A_{ij} \in [0,1]$ and $\triangle p_{ij} \in \mathbb R^{2}$ are generated from the query, indicating the attention weight and sampling offset of the $j^{th}$ sampled point in the $i^{th}$ attention head. $F(p + \triangle p_{ij})$ represents the feature of the sampled point $p + \triangle p_{ij}$ obtained by bilinear interpolation. 

Within our voxel-to-range fusion, we treat occupied voxels as queries to dynamically capture the range-view feature, benefiting from its rich context. The updated voxel-view feature $\mathbf{F}^{3D}_{v2r} \in \mmR^{N_{voxel} \times C}$ is computed by
\begin{equation}
    \label{eq:voxel2range}
    \mathbf{F}^{3D}_{v2r} = \text{DeformAttn}(\mathbf{F}^{3D}, \mathcal{P}(\mathbf{C}), \mathbf{F}^{2D}),
\end{equation}
where $\mathbf{C} \in \mathbb R^{{N}_{voxel} \times 3}$ denotes the corresponding real-world central coordinates of occupied voxels. $\mathcal{P}$ represents the spherical projection function. The updated feature is fed into the feed-forward network, which consists of a linear layer, an activation function, and another linear layer.

\begin{table*}
\centering
\caption{Comparisons on nuScenes-Occupancy validation set. The C, D, and L denote camera, depth, and LiDAR. 
The $\dagger$ indicates the results based on multi-sweep point clouds (typically 10 sweeps). The \textbf{bold} numbers indicate the best results.} 
  \scalebox{0.90}{
\begin{tabular}{l|c|cc|cccccccccccccccc}
\hline
Method 
& \rotatebox{90}{Modality} 
& \rotatebox{90}{IoU (\%)} 
& \rotatebox{90}{mIoU (\%)} 
& \rotatebox{90}{barrier} 
& \rotatebox{90}{bicycle} 
& \rotatebox{90}{bus} 
& \rotatebox{90}{car} 
& \rotatebox{90}{Cons. Veh} 
& \rotatebox{90}{motorcycle} 
& \rotatebox{90}{pedestrian} 
& \rotatebox{90}{traffic cone} 
& \rotatebox{90}{trailer} 
& \rotatebox{90}{truck} 
& \rotatebox{90}{Dri. Sur} 
& \rotatebox{90}{other flat} 
& \rotatebox{90}{sidewalk} 
& \rotatebox{90}{terrain} 
& \rotatebox{90}{manmade} 
& \rotatebox{90}{vegetation} 
\\ 

\hline\hline
MonoScene~\cite{cao2022monoscene} & C & 17.1 & 7.2 & 7.3 & 4.3 & 9.6 & 7.1 & 6.2 & 3.5 & 5.9 & 4.7 & 5.6 & 4.9 & 15.6 & 6.8 & 7.9 & 7.6 & 10.5 & 7.9 \\
TPVFormer~\cite{huang2023tri} & C & 15.1 & 8.3 & 9.7 & 4.5 & 11.5 & 10.7 & 5.5 & 4.6 & 6.3 & 5.4 & 6.9 & 6.9 & 14.1 & 9.8 & 8.9 & 9.0 & 9.9 & 8.5 \\
C-OpenOcc~\cite{wang2023openoccupancy} & C & 17.9 & 10.9 & 9.3 & 7.2 & 11.0 & 12.5 & 7.0 & 9.3 & 8.9 & 5.2 & 4.9 & 10.2 & 23.1 & 17.4 & 15.4 & 14.3 & 8.4 & 11.0 \\
C-CONet~\cite{wang2023openoccupancy} & C & 21.6 & 13.6 & 13.6 & 8.4 & 14.7 & 18.3 & 7.1 & 11.0 & 11.8 & 8.8 & 5.2 & 13.0 & 32.7 & 21.1 & 20.1 & 17.6 & 5.1 & 8.4 \\
SparseOcc~\cite{tang2024sparseocc} & C & 21.8 & 14.1 & 16.1 & 9.3 & 15.1 & 18.6 & 7.3 & 9.4 & 11.2 & 9.4 & 7.2 & 13.0 & 31.8 & 21.7 & 20.7 & 18.8 & 6.1 & 10.6 \\
AICNet~\cite{li2020anisotropic} & C\&D & 23.2 & 10.9 & 11.8 & 4.5 & 12.1 & 12.7 & 6.0 & 3.9 & 6.4 & 6.3 & 8.4 & 7.8 & 24.2 & 13.4 & 13.0 & 11.9 & 11.5 & 20.5 \\
3DSketch~\cite{chen20203d} & C\&D & 25.3 & 11.0 & 12.3 & 5.2 & 10.3 & 12.1 & 7.1 & 4.9 & 5.5 & 6.9 & 8.4 & 7.4 & 21.9 & 15.4 & 13.6 & 12.1 & 12.1 & 21.2 \\
LMSCNet$^\dagger$~\cite{roldao2020lmscnet} & L & 26.7 & 11.8 & 12.9 & 5.2 & 12.8 & 12.6 & 6.6 & 4.9 & 6.3 & 6.5 & 8.8 & 7.7 & 24.3 & 12.7 & 16.5 & 14.5 & 14.2 & 22.1 \\
L-OpenOcc$^\dagger$~\cite{wang2023openoccupancy} & L & 22.3 & 11.9 & 11.1 & 4.0 & 11.4 & 12.9 & 7.2 & 6.2 & 10.1 & 4.4 & 8.1 & 11.0 & 23.3 & 15.8 & 15.5 & 15.6 & 15.0 & 18.7 \\
JS3C-Net$^\dagger$~\cite{yan2021sparse} & L & 29.6 & 12.7 & 14.5 & 4.4 & 13.5 & 12.0 & 7.8 & 4.4 & 7.3 & 6.9 & 9.2 & 9.2 & 27.4 & 15.8 & 15.9 & 16.4 & 14.0 & 24.8 \\
L-CONet$^\dagger$~\cite{wang2023openoccupancy} & L & 30.1 & 15.9 & 18.0 & 3.9 & 14.2 & 18.7 & 8.3 & 6.3 & 11.0 & 5.8 & 14.1 & 14.3 & 35.3 & 20.2 & 21.5 & 20.9 & 19.2 & 23.0 \\ 
L-OccGen$^\dagger$~\cite{wang2024occgen} & L & \textbf{31.6} & 16.8 & 18.8 & 5.1 & 14.8 & 19.6 & 7.0 & 7.7 & 11.5 & 6.7 & 13.9 & 14.6 & \textbf{36.4} & \textbf{22.1} & \textbf{22.8} & 22.3 & 20.6 & 24.5 \\
\hline
\shortname (Ours) & L & 30.6 & \textbf{21.3} & \textbf{22.1} & \textbf{13.5} & \textbf{22.1} & \textbf{24.4} & \textbf{13.2} & \textbf{20.9} & \textbf{25.6} & \textbf{14.5} & \textbf{15.5} & \textbf{20.8} & 31.2 & 19.3 & 22.6 & \textbf{22.8} & \textbf{24.4} & \textbf{27.6} \\
\hline
\end{tabular}
}
\label{tab:nus_occupancy_results}
\end{table*}

\noindent\textbf{Range-to-voxel fusion.}
While the voxel-to-range fusion enables occupied voxels to capture the context of interest from the range-view image, it may suffer from semantic ambiguity, arising from the ``many-to-one" mapping of the 3D-to-2D query mechanism. To alleviate it, we propose leveraging the potential spatial information from the range-view image. Specifically, we use point clouds $\mathbf{P} \in \mathbb R^{N_{point} \times 4}$ as proxies and apply range-to-point transformation~\cite{xu2021rpvnet, xu2023frnet} to transfer the range-view feature $\mathbf{F}^{2D} \in \mathbb R^{C \times H_{f} \times W_{f}}$ into point-view $\mathbf{F}^{3D}_{point} \in \mmR^{N_{point} \times C}$, thus taking advantage of the spatial information of point clouds. Then, we perform 3D voxelization on the point-view feature to generate the voxel-view feature $\mathbf{F}^{3D}_{voxel} \in \mmR^{N_{voxel} \times C}$, which is added to $\mathbf{F}^{3D} \in \mmR^{N_{voxel} \times C}$ for fusion. Lastly, we design the SparseVFE consisting of several sparse basic blocks built on 3D sparse convolution~\cite{spconv2022} and skip connections to further refine and enhance the sparse voxel-view feature. The fused voxel-view feature $\mathbf{F}^{3D}_{r2v} \in \mmR^{N_{voxel} \times C}$ is computed by:
\begin{equation}
    \label{eq:range2voxel}
    \mathbf{F}^{3D}_{r2v} = \text{SparseVFE}(\mathbf{F}^{3D} + \mathbf{F}^{3D}_{voxel}).
\end{equation}

Given the dual-path outputs $\mathbf{F}^{3D}_{v2r}$ and $\mathbf{F}^{3D}_{r2v}$, we concatenate them along the feature dimension, which is fed into 3D sparse convolutions for further fusion. The fused voxel-view feature $\mathbf{F}^{3D}_{fusion} \in \mmR^{N_{voxel} \times C}$ is computed by
\begin{equation}
    \label{eq:fusion}
    \mathbf{F}^{3D}_{fusion} = \text{Fusion}([\mathbf{F}^{3D}_{v2r}, \mathbf{F}^{3D}_{r2v}]),
\end{equation}
where $[\cdot,\cdot]$ denotes the concatenation operation. Finally, we upsample $\mathbf{F}^{3D}_{fusion}$ to the target resolution via coarse-to-fine mapping~\cite{wang2023openoccupancy} and employ the occupancy head to predict the semantic occupancy $\hat{\mathbf{O}}_{sem} \in \mmR^{S \times N^{\prime}_{voxel}}$ of fine voxels.

\subsection{Optimization and Training Details}

To optimize the occupancy network, we use focal loss~\cite{lin2017focal} and dice loss~\cite{sudre2017generalised} as objectives. Specifically, for the predicted geometric occupancy $\hat{\mathbf{O}}_{geo} \in \mmR^{2 \times H \times W \times D}$, we obtain geometric labels $\mathbf{O}_{geo} \in \mmR^{H \times W \times D}$ from original labels. The objective of geometry occupancy prediction is defined as
\begin{equation}
    \label{eq:loss_geo}
    \begin{split}
        \mL_{geo} =& \lambda_{focal}\mL_{focal}(\hat{\bO}_{geo},\bO_{geo}) \\
        &+ \lambda_{dice} \mL_{dice}(\hat{\bO}_{geo},\bO_{geo}).
    \end{split}
\end{equation}

For the predicted semantic occupancy $\hat{\mathbf{O}}_{sem} \in \mmR^{S \times N^{\prime}_{voxel}}$, we obtain the semantic labels $\mathbf{O}_{sem} \in \mathbb R^{N^{\prime}_{voxel}}$ by indexing from original labels. The objective of semantic occupancy prediction is defined as
\begin{equation}
    \label{eq:loss_sem}
    \begin{split}
        \mL_{sem} =& \lambda_{focal}\mL_{focal}(\hat{\bO}_{sem},\bO_{sem}) \\
        &+ \lambda_{dice} \mL_{dice}(\hat{\bO}_{sem},\bO_{sem}).
    \end{split}
\end{equation}

To jointly model geometric and semantic occupancy, we optimize our \shortname using the following objective function:
\begin{equation}
    \label{eq:loss_overall}
    \mathcal L = \mL_{geo} + \mL_{sem}.
\end{equation}

\section{Experiments}
\label{sec:experiments}
 \subsection{Experimental Setup}

\noindent\textbf{Dataset.}
We empirically evaluate our method on nuScenes-Occupancy~\cite{wang2023openoccupancy}, SemanticKITTI~\cite{behley2019semantickitti}, and SemanticPOSS~\cite{pan2020semanticposs}. 
\textbf{nuScenes-Occupancy} is a large-scale outdoor dataset based on nuScenes~\cite{caesar2020nuscenes}, which adopts a Velodyne HDL-32E laser scanner to collect 360° point clouds. The evaluation range for 3D occupancy prediction is $[-51.2m, 51.2m]$ along the X and Y axes and $[-5m, 3m]$ along the Z axis, voxelized into a grid of $512\times512\times40$, with each voxel measuring $0.2m\times0.2m\times0.2m$ and labeled as one of 17 classes (16 semantic, 1 empty). The dataset contains 700 training scenes and 150 validation scenes. 
\textbf{SemanticKITTI} is a semantic scene completion dataset built upon the KITTI Odometry benchmark~\cite{geiger2012we}, using a Velodyne HDL-64E LiDAR sensor to acquire point cloud scans. The scene is defined within a spatial range from $[0m, -25.6m, -2m]$ to $[51.2m, 25.6m, 4.4m]$ along the X, Y, and Z axes, respectively. This volume is discretized into a $256\times256\times32$ voxel grid with a voxel size of $0.2m\times0.2m\times0.2m$. The dataset includes 20 classes (19 semantic, 1 empty) and partitions its 22 sequences as follows: sequences 00–07 and 09–10 for training, sequence 08 for validation, and sequences 11–21 for testing.
\textbf{SemanticPOSS} is another semantic scene completion dataset, captured using a Hesai Pandora 40-line LiDAR sensor. It adopts the same spatial range and voxel grid configuration as SemanticKITTI, spanning from $[0m, -25.6m, -2m]$ to $[51.2m, 25.6m, 4.4m]$ and discretized into a $256\times256\times32$ voxel grid with $0.2m$ resolution. The dataset provides annotations for 12 classes (11 semantic, 1 empty) and comprises six sequences in total, where sequences 00–01 and 03–05 are used for training and sequence 02 is reserved for validation.

\begin{figure}
    \centering
    \includegraphics[width=1.0\linewidth]{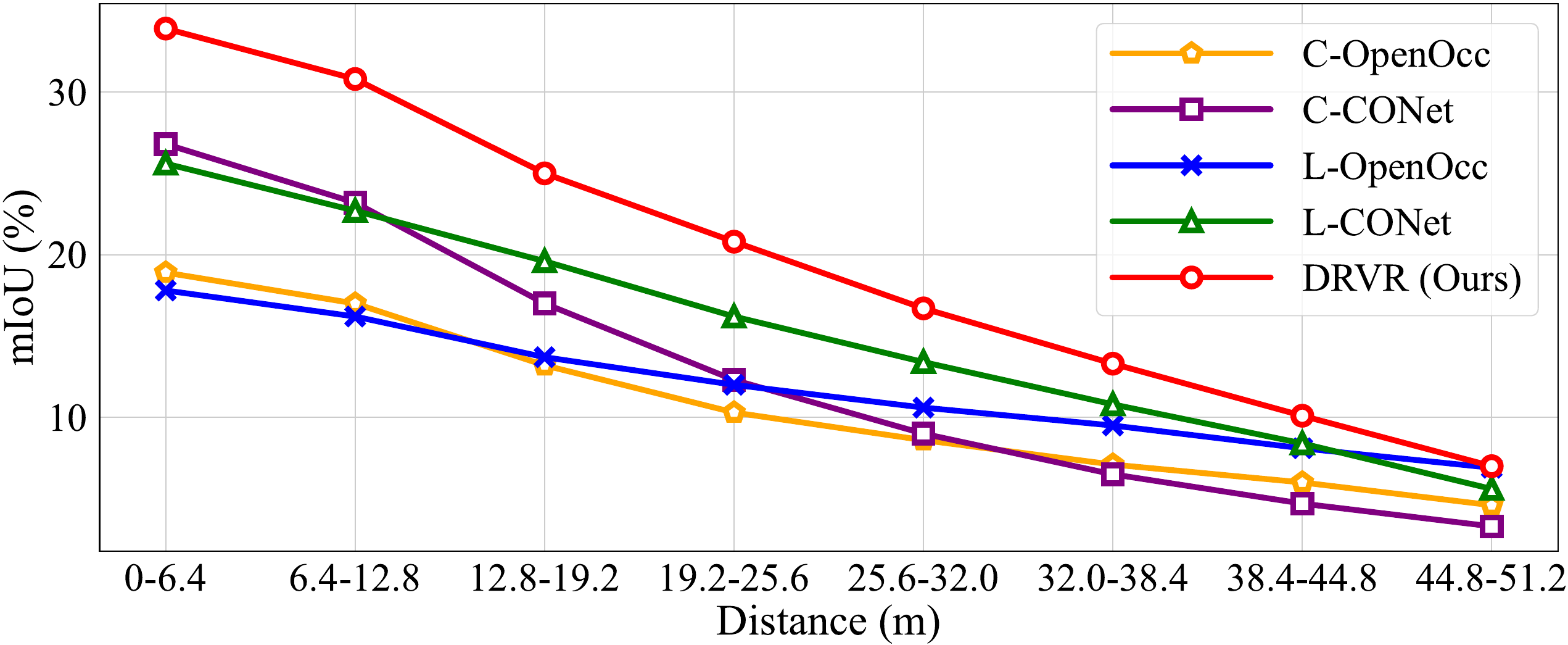}
    \vspace{-0.6cm}
    \caption{Distance-based evaluation on nuScenes-Occupancy validation set.}
    \label{fig:miou_distance_evaluation}
\end{figure}

\begin{table*}[ht]
\centering
\caption{Comparisons on SemanticKITTI test set. The C, L denote camera and LiDAR. The \textbf{bold} numbers indicate the best results.} 
\scalebox{0.82}{
\begin{tabular}{@{}l|c|cc|ccccccccccccccccccc@{}}
\hline
Method 
& \rotatebox{90}{Modality}
& \rotatebox{90}{IoU (\%)} 
& \rotatebox{90}{mIoU (\%)} 
& \rotatebox{90}{car} 
& \rotatebox{90}{bicycle} 
& \rotatebox{90}{motorcycle} 
& \rotatebox{90}{truck} 
& \rotatebox{90}{other-vehicle} 
& \rotatebox{90}{person} 
& \rotatebox{90}{bicyclist} 
& \rotatebox{90}{motorcyclist} 
& \rotatebox{90}{road} 
& \rotatebox{90}{parking} 
& \rotatebox{90}{sidewalk} 
& \rotatebox{90}{other-ground} 
& \rotatebox{90}{building} 
& \rotatebox{90}{fence} 
& \rotatebox{90}{vegetation} 
& \rotatebox{90}{trunk} 
& \rotatebox{90}{terrain} 
& \rotatebox{90}{pole} 
& \rotatebox{90}{traffic-sign} 
\\ 

\hline\hline
MonoScene~\cite{cao2022monoscene} & C & 34.2 & 11.1 & 18.8 & 0.5 & 0.7 & 3.3 & 4.4 & 1.0 & 1.4 & 0.4 & 54.7 & 24.8 & 27.1 & 5.7 & 14.4 & 11.1 & 14.9 & 2.4 & 19.5 & 3.3 & 2.1 \\
TPVFormer~\cite{huang2023tri} & C & 34.3 & 11.3 & 19.2 & 1.0 & 0.5 & 3.7 & 2.3 & 1.1 & 2.4 & 0.3 & 55.1 & 27.4 & 27.2 & 6.5 & 14.8 & 11.0 & 13.9 & 2.6 & 20.4 & 2.9 & 1.5 \\
VoxFormer~\cite{li2023voxformer} & C & 43.0 & 12.2 & 20.8 & 1.0 & 0.7 & 3.5 & 3.7 & 1.4 & 2.6 & 0.2 & 53.9 & 21.1 & 25.3 & 5.6 & 19.8 & 11.1 & 22.4 & 7.5 & 21.3 & 5.1 & 4.9 \\
OccFormer~\cite{zhang2023occformer} & C & 34.5 & 12.3 & 21.6 & 1.5 & 1.7 & 1.2 & 3.2 & 2.2 & 1.1 & 0.2 & 55.9 & 31.5 & 30.3 & 6.5 & 15.7 & 11.9 & 16.8 & 3.9 & 21.3 & 3.8 & 3.7 \\
SAG-IRT~\cite{xiao2024semantic} & C & 38.4 & 13.4 & 22.9 & 1.7 & 1.9 & 4.7 & 5.9 & 2.2 & 4.1 & 0.5 & 56.2 & 28.4 & 30.2 & 6.8 & 18.5 & 13.3 & 20.1 & 8.6 & 19.1 & 4.2 & 5.8 \\ 
LinkOcc~\cite{ouyang2024linkocc} & C & 43.0 & 13.8 & 23.2 & 0.3 & 2.0 & 5.1 & 5.0 & 2.2 & 2.0 & 0.0 & 55.0 & 25.4 & 28.5 & 7.1 & 24.1 & 13.7 & 23.2 & 8.1 & 27.0 & 7.8 & 6.0 \\
Symphonies~\cite{jiang2024symphonize} & C & 42.2 & 15.0 & 23.6 & 3.6 & 2.6 & 3.2 & 5.6 & 3.2 & 1.9 & 2.0 & 58.4 & 26.9 & 29.3 & 11.7 & 24.7 & 16.1 & 24.2 & 10.0 & 23.1 & 7.7 & 8.0 \\
CGFormer~\cite{yu2024context} & C & 44.4 & 16.6 & 26.1 & 3.7 & 1.3 & 4.3 & 2.7 & 1.7 & 3.6 & 0.4 & 64.3 & 34.1 & 34.2 & 12.1 & 25.8 & 18.7 & 24.5 & 11.2 & 29.3 & 8.7 & 9.3 \\
VisHall3D\cite{lu2025vishall3d} & C & 46.5 & 17.5 & 26.7 & 2.9 & 3.3 & 7.5 & 6.2 & 2.3 & 5.1 & 1.9 & 64.6 & 32.0 & 34.1 & 12.5 & 26.9 & 19.5 & 27.3 & 12.5 & 28.0 & 9.2 & 9.2 \\
SSCNet-Full~\cite{song2017semantic} & L & 50.0 & 16.1 & 24.3 & 0.5 & 0.8 & 1.2 & 4.3 & 0.3 & 0.3 & 0.0 & 51.2 & 27.1 & 30.8 & 6.4 & 34.5 & 19.9 & 35.3 & 18.2 & 29.0 & 13.1 & 6.7 \\
ESSCNet~\cite{zhang2018efficient} & L & 41.8 & 17.5 & 26.4 & 0.3 & 5.4 & 5.0 & 9.1 & 2.9 & 2.7 & 0.1 & 43.8 & 26.9 & 28.1 & 10.3 & 29.8 & 23.3 & 35.8 & 20.1 & 28.7 & 16.4 & 16.7 \\
LMSCNet-SS~\cite{roldao2020lmscnet} & L & 56.7 & 17.6 & 30.9 & 0.0 & 0.0 & 1.5 & 0.8 & 0.0 & 0.0 & 0.0 & 64.8 & 29.0 & 34.7 & 4.6 & 38.1 & 21.3 & 41.3 & 19.9 & 32.1 & 15.0 & 0.8 \\
TDS~\cite{garbade2019two} & L & 50.6 & 17.7 & 29.5 & 0.0 & 0.0 & 2.5 & 0.1 & 0.0 & 0.0 & 0.0 & 62.2 & 23.3 & 31.6 & 6.5 & 34.1 & 24.1 & 40.1 & 21.9 & 33.1 & 16.9 & 6.9 \\
UDNet~\cite{zou2021up} & L & 59.4 & 19.5 & 33.9 & 0.8 & 0.4 & 3.8 & 4.4 & 0.5 & 0.3 & 0.3 & 62.0 & 28.2 & 35.1 & 9.1 & 39.5 & 24.4 & 40.9 & 23.2 & 32.3 & 18.8 & 13.1 \\
Local-DIFs~\cite{rist2021semantic} & L & 57.7 & 22.7 & 34.8 & 3.6 & 2.4 & 4.4 & 4.8 & 2.5 & 1.1 & 0.0 & 67.9 & 40.1 & 42.9 & 11.4 & 40.4 & 29.0 & 42.2 & 26.5 & 39.1 & 21.3 & 17.5 \\
SSA-SC~\cite{yang2021semantic} & L & 58.8 & 23.5 & 36.5 & 13.9 & 4.6 & 5.7 & 7.4 & 4.4 & 2.6 & 0.7 & 72.2 & 37.4 & 43.7 & 10.9 & 43.6 & 30.7 & 43.5 & 25.6 & 41.8 & 14.5 & 6.9 \\
JS3C-Net~\cite{yan2021sparse} & L & 56.6 & 23.8 & 33.3 & 14.4 & 8.8 & 7.2 & 12.7 & \textbf{8.0} & \textbf{5.1} & 0.4 & 64.7 & 34.9 & 39.9 & 14.1 & 39.4 & 30.4 & 43.1 & 19.6 & 40.5 & 18.9 & 15.9 \\
SSC-RS~\cite{mei2023ssc} & L & 59.7 & 24.2 & 36.4 & 10.1 & 5.1 & 5.3 & 11.2 & 4.7 & 2.4 & 0.9 & \textbf{73.1} & 38.6 & 44.4 & \textbf{17.4} & \textbf{44.6} & 30.8 & 44.1 & 26.0 & 41.9 & 15.0 & 7.2 \\
VPNet~\cite{wang2024voxel} & L & 60.4 & 25.0 & 37.2 & 14.0 & 9.8 & 4.3 & 8.2 & 4.9 & 2.0 & \textbf{2.4} & 72.4 & 40.5 & 44.3 & 14.8 & 44.0 & 32.7 & 45.3 & 30.9 & 42.1 & 17.1 & 8.8 \\
\hline
\shortname (Ours) & L & \textbf{61.4} & \textbf{28.5} & \textbf{39.6} & \textbf{16.1} & \textbf{13.1} & \textbf{8.7} & \textbf{13.4} & 5.5 & 4.5 & 1.5 & 71.4 & \textbf{44.3} & \textbf{46.9} & 15.8 & 43.9 & \textbf{36.1} & \textbf{47.6} & \textbf{32.9} & \textbf{44.9} & \textbf{30.2} & \textbf{24.3} \\
\hline
\end{tabular}
}
\label{tab:semantic_kitti_results}
\end{table*}

\begin{table*}[ht]
\centering
\caption{Comparisons on SemanticKITTI validation set. The C, L denote camera and LiDAR. The \textbf{bold} numbers indicate the best results.} 
\scalebox{0.80}{
\begin{tabular}{@{}l|c|cc|ccccccccccccccccccc@{}}
\hline
Method 
& \rotatebox{90}{Modality}
& \rotatebox{90}{IoU (\%)} 
& \rotatebox{90}{mIoU (\%)} 
& \rotatebox{90}{car} 
& \rotatebox{90}{bicycle} 
& \rotatebox{90}{motorcycle} 
& \rotatebox{90}{truck} 
& \rotatebox{90}{other-vehicle} 
& \rotatebox{90}{person} 
& \rotatebox{90}{bicyclist} 
& \rotatebox{90}{motorcyclist} 
& \rotatebox{90}{road} 
& \rotatebox{90}{parking} 
& \rotatebox{90}{sidewalk} 
& \rotatebox{90}{other-ground} 
& \rotatebox{90}{building} 
& \rotatebox{90}{fence} 
& \rotatebox{90}{vegetation} 
& \rotatebox{90}{trunk} 
& \rotatebox{90}{terrain} 
& \rotatebox{90}{pole} 
& \rotatebox{90}{traffic-sign} 
\\ 

\hline\hline
MonoScene~\cite{cao2022monoscene} & C & 37.1 & 11.5 & 23.6 & 0.2 & 0.8 & 7.8 & 3.6 & 1.8 & 1.0 & 0.0 & 57.5 & 15.7 & 27.1 & 0.9 & 14.2 & 6.4 & 18.1 & 2.6 & 30.8 & 4.1 & 2.5 \\
TPVFormer~\cite{huang2023tri} & C & 35.6 & 11.4 & 23.8 & 0.4 & 0.1 & 8.1 & 4.4 & 0.5 & 0.9 & 0.0 & 56.5 & 20.6 & 25.9 & 0.9 & 13.9 & 5.9 & 16.9 & 2.3 & 30.4 & 3.1 & 1.5 \\
VoxFormer~\cite{li2023voxformer} & C & 44.0 & 12.4 & 25.8 & 0.6 & 0.5 & 5.6 & 3.8 & 1.8 & 3.3 & 0.0 & 54.8 & 15.5 & 26.4 & 0.7 & 17.7 & 7.6 & 24.4 & 5.1 & 30.0 & 7.1 & 4.2 \\
OccFormer~\cite{zhang2023occformer} & C & 36.5 & 13.5 & 25.1 & 0.8 & 1.2 & 25.5 & 8.5 & 2.8 & 2.8 & 0.0 & 58.9 & 19.6 & 26.9 & 0.3 & 14.4  & 5.6 & 19.6 & 3.9 & 32.6 & 4.3 & 2.9 \\
SAG-IRT~\cite{xiao2024semantic} & C & 40.5 & 13.8 & 27.7 & 1.5 & 2.3 & 16.1 & 9.6 & 3.7 & \textbf{8.6} & 0.0 & 59.3 & 16.2 & 27.8 & 1.1 & 15.9 & 7.8 & 21.0 & 6.6 & 28.7 & 5.1 & 2.8 \\
LinkOcc~\cite{ouyang2024linkocc} & C & 42.1 & 13.9 & 28.8 & 2.3 & 1.5 & 18.0 & 10.1 & 3.8 & 5.1 & 0.0 & 55.1 & 22.6 & 27.9 & 0.5 & 20.3 & 6.0 & 24.0 & 6.9 & 32.9 & 7.3 & 4.0 \\
Symphonies~\cite{jiang2024symphonize} & C & 41.9 & 14.9 & 28.7 & 2.5 & 2.8 & 20.4 & 13.9 & 3.5 & 2.2 & 0.0 & 56.4 & 15.3 & 27.6 & 1.0 & 21.6 & 8.4 & 25.7 & 6.6 & 30.9 & 9.6 & 5.8 \\
CGFormer~\cite{yu2024context} & C & 46.0 & 16.9 & 34.3 & 4.6 & 2.7 & 19.4 & 7.7 & 2.4 & 4.1 & 0.0 & 65.5 & 20.8 & 32.3 & 0.2 & 23.5 & 9.2 & 26.9 & 8.8 & 39.5 & 10.7 & 7.8 \\
VisHall3D\cite{lu2025vishall3d} & C & 46.1 & 17.1 & - & - & - & - & - & - & - & - & - & - & - & - & - & - & - & - & - & - & - \\
LMSCNet-SS~\cite{roldao2020lmscnet} & L & 54.2 & 16.8 & - & - & - & - & - & - & - & - & - & - & - & - & - & - & - & - & - & - & - \\
UDNet~\cite{zou2021up} & L & 58.9 & 20.7 & 42.1 & 1.8 & 2.3 & 25.7 & 11.2 & 2.5 & 1.2 & 0.0 & 67.0 & 20.3 & 37.2 & 2.2 & 36.0 & 11.9 & 40.1 & 18.3 & 45.8 & 23.0 & 3.8 \\
Local-DIFs~\cite{rist2021semantic} & L & 57.6 & 24.0 & - & - & - & - & - & - & - & - & - & - & - & - & - & - & - & - & - & - & - \\
JS3C-Net~\cite{yan2021sparse} & L & 57.0 & 24.0 & - & - & - & - & - & - & - & - & - & - & - & - & - & - & - & - & - & - & - \\
SSA-SC~\cite{yang2021semantic} & L & 58.3 & 24.5 & 47.0 & 9.2 & 7.4 & 39.6 & 19.1 & 6.3 & 3.2 & 0.0 & 72.8 & 21.1 & 44.3 & \textbf{4.1} & 41.5 & 15.2 & 41.9 & 22.0 & 49.4 & 17.8 & 4.4 \\
SSC-RS~\cite{mei2023ssc} & L & 58.6 & 24.8 & 46.8 & 1.5 & 6.9 & 41.5 & 19.8 & 6.2 & 1.5 & 0.0 & 73.8 & 26.6 & 45.3 & 2.1 & 41.0 & 15.8 & 42.6 & 22.2 & \textbf{50.6} & 17.9 & 4.6 \\
VPNet~\cite{wang2024voxel} & L & 59.6 & 26.1 & - & - & - & - & - & - & - & - & - & - & - & - & - & - & - & - & - & - & - \\
OCC-Exoskeleton~\cite{wang2025occ} & L & 59.7 & 26.9 & 47.2 & 13.4 & 17.1 & 34.8 & 20.0 & \textbf{8.0} & 1.9 & 0.0 & \textbf{74.0} & 25.4 & 46.3 & 1.9 & \textbf{41.7} & 16.8 & 43.3 & 23.9 & 50.1 & 29.2 & 15.3 \\
\hline
\shortname (Ours) & L & \textbf{60.3} & \textbf{30.4} & \textbf{51.0} & \textbf{14.1} & \textbf{17.2} & \textbf{61.4} & \textbf{31.6} & 7.9 & 3.3 & 0.0 & 73.5 & \textbf{38.8} & \textbf{49.4} & 0.1 & 41.1 & \textbf{20.5} & \textbf{43.9} & \textbf{27.0} & 47.1 & \textbf{32.6} & \textbf{17.1} \\
\hline
\end{tabular}
}
\label{tab:semantic_kitti_val_results}
\end{table*}

\noindent\textbf{Evaluation metrics.}
Following the common practices~\cite{behley2019semantickitti, wang2023openoccupancy}, we employ the intersection over union (IoU) as the geometric metric, distinguishing between occupied and empty voxels. Specifically:
\begin{equation}
    \mathrm{IoU}=\frac{\mathrm{TP}_{\mathrm{o}}}{\mathrm{TP}_{\mathrm{o}}+\mathrm{FP}_{\mathrm{o}}+\mathrm{FN}_{\mathrm{o}}},
\end{equation}
where $\mathrm{TP}_\mathrm{o}$, $\mathrm{FP}_\mathrm{o}$, $\mathrm{FN}_\mathrm{o}$ are the number of true positive, false positive and false negative predictions for occupied voxels. Moreover, we report the mean intersection over union (mIoU) of all semantic classes to evaluate the quality of semantic segmentation on the occupied voxels:
\begin{equation}
    \mathrm{mIoU}=\frac{1}{\mathrm{C}}\sum\limits_{\mathrm{c}=1}^{\mathrm{C}}\frac{\mathrm{TP}_{\mathrm{c}}}{\mathrm{TP}_{\mathrm{c}}+\mathrm{FP}_{\mathrm{c}}+\mathrm{FN}_{\mathrm{c}}},
\end{equation}
where $\mathrm{TP}_\mathrm{c}$, $\mathrm{FP}_\mathrm{c}$, $\mathrm{FN}_\mathrm{c}$ are the number of true positive, false positive and false negative predictions for class $\mathrm{c}$, and $\mathrm{C}$ denotes the number of semantic classes.

\begin{table*}[ht]
\centering
\caption{Comparisons on SemanticPOSS validation set. The L denotes LiDAR. The \textbf{bold} numbers indicate the best results.} 
\scalebox{1.24}{
\begin{tabular}{@{}l|c|cc|ccccccccccc@{}}
\hline
Method 
& \rotatebox{90}{Modality}
& \rotatebox{90}{IoU (\%)} 
& \rotatebox{90}{mIoU (\%)} 
& \rotatebox{90}{person} 
& \rotatebox{90}{rider} 
& \rotatebox{90}{car} 
& \rotatebox{90}{trunk} 
& \rotatebox{90}{plants} 
& \rotatebox{90}{traffic-sign} 
& \rotatebox{90}{pole} 
& \rotatebox{90}{building} 
& \rotatebox{90}{fence} 
& \rotatebox{90}{bike} 
& \rotatebox{90}{ground}
\\ 

\hline\hline
SSCNet-Full~\cite{song2017semantic} & L & 41.7 & 12.1 & 8.2 & 0.4 & 1.3 & 3.7 & 34.4 & 1.0 & 4.8 & 34.3 & 3.3 & 16.0 & 25.6  \\
LMSCNet-SS~\cite{roldao2020lmscnet} & L & 52.6 & 16.4 & 8.4 & 0.0 & 1.0 & 4.3 & 3.8 & 1.8 & 0.8 & 39.2 & 13.6 & 27.4 & 45.4  \\
SSA-SC~\cite{yang2021semantic} & L & 53.3 & 21.6 & \textbf{17.8} & 0.5 & 4.3 & 3.1 & 43.3 & 2.6 & \textbf{11.1} & 40.6 & 23.4 & 42.5 & 48.2  \\
VPNet~\cite{wang2024voxel} & L & 56.9 & 22.4 & 15.1 & 0.4 & 1.7 & 4.9 & 46.4 & 1.0 & 9.4 & \textbf{43.2} & \textbf{28.0} & 45.0 & 51.0 \\
\hline
\shortname (Ours) & L & \textbf{58.4} & \textbf{22.7} & 11.4 & \textbf{0.6} & \textbf{5.5} & \textbf{5.0} & \textbf{46.8} & \textbf{3.6} & 10.0 & 42.0 & 27.1 & \textbf{45.1} & \textbf{52.1}  \\
\hline
\end{tabular}
}
\label{tab:semantic_poss_results}
\end{table*}

\begin{figure*}[ht]
    \centering
    \includegraphics[width=1.0\linewidth]{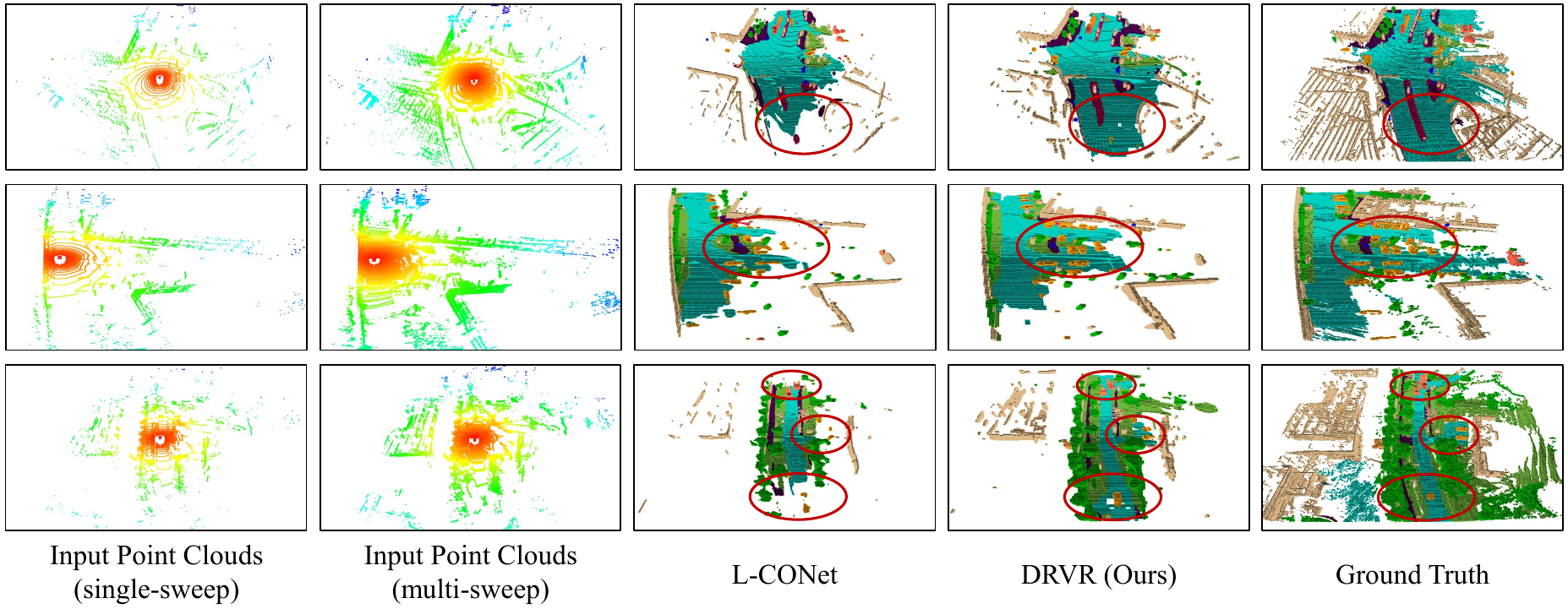}
    \vspace{-0.6cm}
    \caption{Qualitative results of \shortname on nuScenes-Occupancy. We highlight the main differences with red circles. Better viewed by zooming in.}
    \label{fig:nus_visualize}
\end{figure*}

\begin{figure*}[ht]
    \centering
    \includegraphics[width=1.0\linewidth]{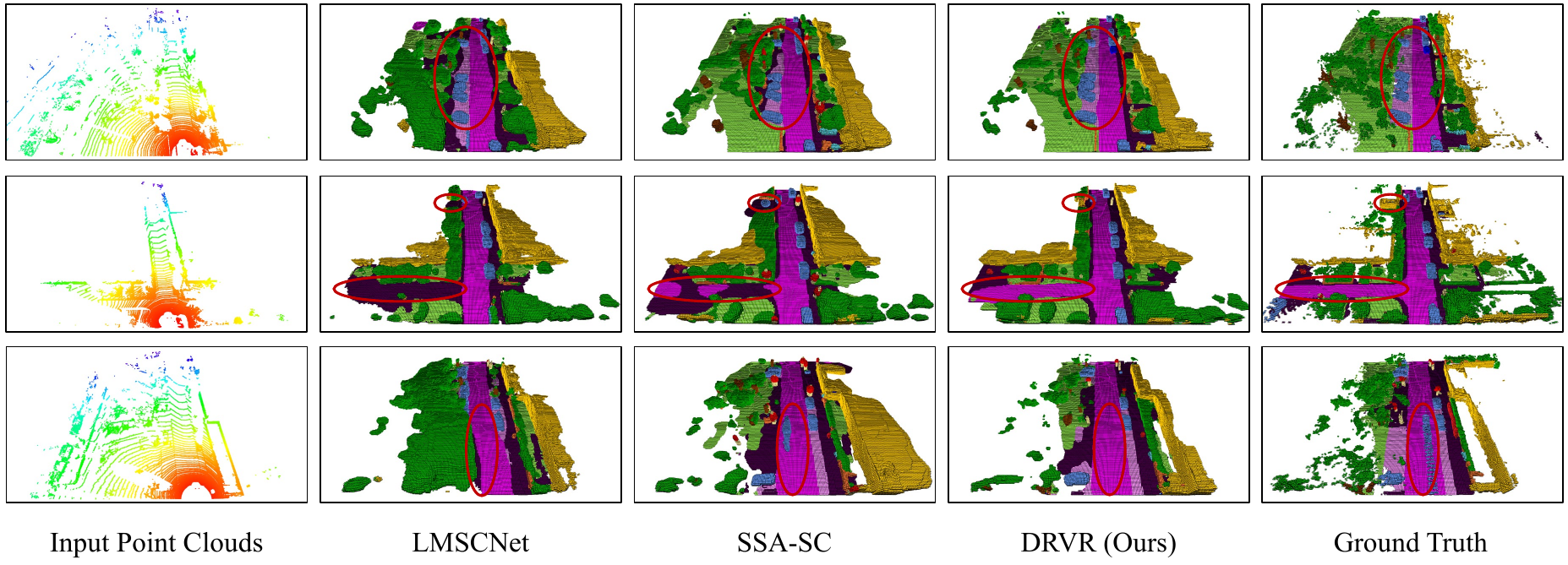}
    \vspace{-0.6cm}
    \caption{Qualitative results of \shortname on SemanticKITTI. We highlight the main differences with red circles. Better viewed by zooming in.}
    \label{fig:smk_visualize}
\end{figure*}

\noindent\textbf{Implementation details.}
We implement our method using PyTorch~\cite{paszke2019pytorch} and apply pre-trained CENet~\cite{cheng2022cenet} as our range-view encoder. The number of voxel sizes for the voxel-view encoder is set to 3, where the voxel resolutions are $128\times128\times16$, $64\times64\times8$, and $32\times32\times4$. The voxel-to-range fusion consists of 3 deformable attention layers, each with 4 attention heads and 8 sampled points around reference points. The range-to-voxel fusion contains 4 sparse basic blocks with skip connections. The feature dimension $C$ is 128 for three benchmarks. The sizes of the range-view image are $64\times2048$ for nuScenes-Occupancy and $128\times1024$ for SemanticKITTI and SemanticPOSS, respectively. The spatial resolutions of geometry occupancy and semantic occupancy are $128\times128\times10$, $256\times256\times20$ for nuScenes-Occupancy and $128\times128\times16$, $256\times256\times32$ for SemanticKITTI and SemanticPOSS. We train the model using the AdamW~\cite{loshchilov2017decoupled} optimizer for 50 epochs on nuScenes-Occupancy and 30 epochs on both SemanticKITTI and SemanticPOSS, with a weight decay of 0.01. The initial learning rates are 0.001 for nuScenes-Occupancy and 0.0001 for both SemanticKITTI and SemanticPOSS, decaying to 0 following a cosine policy. We conduct experiments on nuScenes-Occupancy with a batch size of 8 using 8 NVIDIA 3090 GPUs. On SemanticKITTI and SemanticPOSS, we use 4 GPUs with a batch size of 4. $\lambda_{focal}$ and $\lambda_{dice}$ are set to 1.0 on three datasets. During inference on nuScenes-Occupancy, we generate the target semantic occupancy representation with the resolution of $512\times512\times40$ by performing coarse-to-fine mapping~\cite{wang2023openoccupancy} on the model output. All ablation studies are conducted on the SemanticKITTI validation set.

\subsection{Benchmark Results}

\noindent\textbf{Quantitative comparisons on nuScenes-Occupancy.}
We evaluate our method on the nuScenes-Occupancy validation set. As shown in Table~\ref{tab:nus_occupancy_results}, we report both the geometry metric IoU and the semantic metric mIoU. From the results, our \shortname achieves an IoU of 30.6\% and a mIoU of 21.3\%, demonstrating its superior performance. Specifically, our method significantly outperforms all camera-based methods in terms of IoU and mIoU. Compared to LiDAR-based methods LMSCNet~\cite{roldao2020lmscnet}, JS3C-Net~\cite{yan2021sparse}, L-CONet~\cite{wang2023openoccupancy} and L-OccGen~\cite{wang2024occgen}, which stack multi-sweep point clouds for dense spatial information, our single-sweep \shortname surpasses them by 9.5\%, 8.6\%, 5.4\% and 4.5\% in mIoU, indicating our method can effectively reduce dependence on high-resolution LiDAR while maintaining strong performance. For small objects (\textit{e.g.}, bicycle, car, motorcycle, pedestrian), our \shortname exceeds other methods by a large margin, showing our method is capable of learning semantic representation even under sparse observation conditions. Moreover, we conduct a distance-based evaluation to analyze the perception capability of our method across different ranges. As shown in Figure~\ref{fig:miou_distance_evaluation}, our \shortname consistently achieves the best performance at various distances.

\noindent\textbf{Quantitative comparisons on SemanticKITTI.}
To verify the generalization of the proposed method on dense point clouds, we evaluate our method on SemanticKITTI. Since SemanticKITTI is captured using a 64-line LiDAR, it provides significantly denser point clouds compared to nuScenes-Occupancy (32-line LiDAR). Therefore, existing methods on SemanticKITTI can use single-sweep point clouds as input. However, such high-resolution LiDAR sensors are considerably more expensive than their low-resolution counterparts. In contrast to these methods that rely on dense point clouds from costly high-resolution LiDARs, our \shortname is primarily designed for sparse point clouds yet remains readily adaptable to dense scenarios, demonstrating its flexibility and cost-effectiveness. As shown in Table~\ref{tab:semantic_kitti_results}, our \shortname achieves the best performance (61.4\% IoU, 28.5\% mIoU) on the SemanticKITTI test set. Specifically, our \shortname outperforms SSA-SC~\cite{yang2021semantic}, JS3C-Net~\cite{yan2021sparse}, SSC-RS~\cite{mei2023ssc} and VPNet~\cite{wang2024voxel} by 2.6\%, 4.8\%, 1.7\%, 1.0\% in IoU, and by 5.0\%, 4.7\%, 4.3\%, 3.5\% in mIoU, respectively. Moreover, we report the results on the SemanticKITTI validation set. As shown in Table~\ref{tab:semantic_kitti_val_results}, our \shortname surpasses camera-based and LiDAR-based methods in terms of IoU and mIoU by a large margin, demonstrating the superior performance of our method.

\noindent\textbf{Quantitative comparisons on SemanticPOSS.}
Following~\cite{wang2024voxel}, we further evaluate our method on SemanticPOSS and compare it against existing methods, including SSCNet~\cite{song2017semantic}, LMSCNet~\cite{roldao2020lmscnet}, SSA-SC~\cite{yang2021semantic}, and VPNet~\cite{wang2024voxel}. As shown in Table~\ref{tab:semantic_poss_results}, our \shortname outperforms existing methods in both geometry IoU and semantic mIoU on the SemanticPOSS validation set. Specifically, our \shortname surpasses SSA-SC~\cite{yang2021semantic}, and VPNet~\cite{wang2024voxel} by 5.1\%, 1,5\% in IoU and 1.1\%, 0.3\% in mIoU.

\begin{table}
    \centering
    \caption{Comparisons of model efficiency on nuScenes-Occupancy validation set. The \textbf{bold} numbers indicate the best results.}
    \scalebox{0.86}{
        \begin{tabular}{@{}l|c|cccc@{}}
        \hline
        Method & Modality & \#Params. & Inference time & IoU (\%) & mIoU (\%) \\
        \hline
        C-OpenOcc~\cite{wang2023openoccupancy} & Camera & 99.20M & 274.09ms & 17.9 & 10.9 \\
        C-CONet~\cite{wang2023openoccupancy} & Camera & 117.99M & 394.77ms & 21.6 & 13.6 \\
        L-OpenOcc~\cite{wang2023openoccupancy} & LiDAR & 65.62M & 229.98ms & 22.3 & 11.9 \\
        L-CONet~\cite{wang2023openoccupancy} & LiDAR & 65.63M & 266.32ms & 30.1 & 15.9 \\
        \hline
        \shortname(Ours) & LiDAR & \textbf{19.67M} & \textbf{128.03ms} & \textbf{30.6} & \textbf{21.3} \\
        \hline
        \end{tabular}
    }
    \label{tab:efficiency_analysis}
\end{table}

\begin{table}
    \centering
    \caption{Ablation study of the proposed model components on SemanticKITTI validation set. The \textbf{bold} numbers denote the best results.}
    \scalebox{1.2}{
        \begin{tabular}{@{}cl|cc@{}}
        \hline
        & Method & IoU (\%) & mIoU (\%) \\
        \hline
        1 & w/o Range-view Encoder & 58.6 & 25.3 \\
        2 & w/o Voxel-view Encoder & 55.0 & 24.8 \\
        3 & w/o Range-to-Voxel Fusion & 59.6 & 29.6 \\
        4 & w/o Voxel-to-Range Fusion & 59.7 & 30.0 \\
        \hline
        5 & \shortname (Ours) & \textbf{60.3} & \textbf{30.4} \\
        \hline
        \end{tabular}
        }
    \label{tab:model_components}
\end{table}

\subsection{Qualitative Evaluation}
To better understand the benefits of our method, we provide qualitative evaluations on nuScenes-Occupancy. As shown in Figure~\ref{fig:nus_visualize}, multi-sweep point clouds effectively capture the surrounding scene, whereas single-sweep inputs suffer from severe sparsity, making scene understanding particularly challenging. Remarkably, despite using single-sweep sparse point clouds as input, our \shortname outperforms the multi-sweep L-CONet~\cite{wang2023openoccupancy}, demonstrating the effectiveness of our method. For example, our method accurately recovers fine-grained structures (\eg, small objects) and reconstructs background regions more completely and precisely, aligning well with the quantitative results.
Moreover, we visualize semantic occupancy predictions on SemanticKITTI in Figure~\ref{fig:smk_visualize}. Compared to LMSCNet~\cite{roldao2020lmscnet} and SSA-SC~\cite{yang2021semantic}, our method exhibits superior scene completion performance across both short- and long-range regions. We observe that the ground truth may contain noise, as it is constructed by aggregating multi-sweep point clouds. In this case, our method shows stronger robustness to noisy labels compared to SSA-SC~\cite{yang2021semantic}.

\begin{table}
    \centering
    \caption{Effect of using different resolutions of voxel-view representation on SemanticKITTI validation set. The \textbf{bold} numbers denote the best results.}
    \scalebox{1.2}{
        \begin{tabular}{@{}cccc|cc@{}}
        \hline
        & \multicolumn{3}{c|}{Voxel Resolution} & \multirow{2}{*}{IoU (\%)} & \multirow{2}{*}{mIoU (\%)} \\
        & 32 & 64 & 128 & & \\
        \hline
        1 & $\checkmark$ & & & 58.0 & 28.8 \\
        2 & & $\checkmark$ & & 58.0 & 28.2 \\
        3 & & & $\checkmark$ & 56.4 & 28.3 \\
        \hline
        4 & $\checkmark$ & $\checkmark$ & $\checkmark$ & \textbf{60.3} & \textbf{30.4} \\
        \hline
        \end{tabular}
    }
    \label{tab:voxel_sizes}
\end{table}

\begin{table}[t]
    \centering
    \caption{Effect of using different stages of range-view representation on SemanticKITTI validation set. The \textbf{bold} numbers denote the best results.}
    \scalebox{0.90}{
    \begin{tabular}{@{}cccccc|cc@{}}
    \hline
    & Stage-1 & Stage-2 & Stage-3 & Stage-4 & Stage-5 & IoU (\%) & mIoU (\%) \\
    \hline
    1 & $\checkmark$ & & & & & 59.7 & 26.7 \\
    2 & & $\checkmark$ & & & & 59.8 & 28.3 \\
    3 & & & $\checkmark$ & & & 59.8 & 29.1 \\
    4 & & & & $\checkmark$ & & 60.1 & 29.5 \\
    5 & & & & & $\checkmark$ & 59.6 & 29.5 \\
    \hline
    6 & $\checkmark$ & $\checkmark$ & $\checkmark$ & $\checkmark$ & $\checkmark$ & \textbf{60.3} & \textbf{30.4} \\
    \hline
    \end{tabular}
    }
    \label{tab:range_layers}
\end{table}

\subsection{Efficiency Analysis}
We evaluate the efficiency of our method on the nuScenes-Occupancy validation set with a single GeForce RTX 3090. Our \shortname leverages the range-view context and voxel-view geometry of single-sweep sparse point clouds to generate 3D semantic occupancy representation, which alleviates the dependency on the multi-sweeps and speeds up the inference. As shown in Table~\ref{tab:efficiency_analysis}, our \shortname achieves better performance with fewer model parameters and maintains an acceptable runtime compared with other methods. Notably, our \shortname is $2.1\times$ (128.03ms vs. 266.32ms) faster than L-CONet~\cite{wang2023openoccupancy} with 5.4\% improvements in semantic mIoU.

\subsection{Ablation Study}
\label{ablation_study}
\noindent\textbf{Effect of the proposed model components.}
We study the effect of the proposed model components of our \shortname on the SemanticKITTI validation set, including context-aware range-view encoder, geometry-aware voxel-view encoder, range-to-voxel fusion, and voxel-to-range fusion. 
Specifically, we ablate either the range-view or the voxel-view encoder to simulate single-view baselines. When the range-view encoder is removed, the model relies solely on the voxel-view encoder, which processes the voxel-view geometry using several sparse basic blocks to predict semantic occupancy. Conversely, when the voxel-view encoder is omitted, we adopt the paradigm of existing camera-based methods~\cite{li2023voxformer, jiang2024symphonize}, representing the 3D scene as a dense grid of learnable queries that interact with range-view context via deformable cross-attention~\cite{zhu2020deformable}. Moreover, we further evaluate the fusion module by removing either the voxel-to-range or the range-to-voxel interaction pathway.
The experimental results are shown in Table~\ref{tab:model_components}. Comparing the first and fifth lines, the range-view encoder brings 5.1\% (30.4\% vs. 25.3\%) mIoU and 1.7\% (60.3\% vs. 58.6\%) IoU improvements, indicating the effectiveness of rich context in range-view representation. From the second and fifth lines, the voxel-view encoder boosts the model's performance by 5.6\% in mIoU (30.4\% vs. 24.8\%) and 5.3\% in IoU (60.3\% vs. 55.0\%), demonstrating the role of geometric information of multi-scale voxel-view representations. The remaining lines show that the range-to-voxel and voxel-to-range fusions promote the model's performance by 0.8\%, 0.4\% in semantic mIoU and 0.7\%, 0.6\% in geometry IoU, respectively, underscoring the value of bidirectional fusion.

\begin{table}
    \centering
    \caption{Effect of the number of voxel-to-range fusion modules on SemanticKITTI validation set. The \textbf{bold} numbers denote the best results.}
    \scalebox{1.2}{
        \begin{tabular}{@{}c|ccccc@{}}
        \hline
        Number & 1 & 2 & 3 & 4 & 5 \\
        \hline
        IoU (\%) & 59.8 & 59.8 & 60.3 & 60.3 & \textbf{60.5} \\
        mIoU (\%) & 30.1 & 30.0 & \textbf{30.4} & 30.2 & \textbf{30.4} \\
        \hline
        \end{tabular}
        }
    \label{tab:num_deformattn}
\end{table}

\begin{table}
    \centering
    \caption{Effect of the depth of range-to-voxel fusion module on SemanticKITTI validation set. The \textbf{bold} numbers denote the best results.}
    \scalebox{1.2}{
        \begin{tabular}{@{}c|ccccc@{}}
        \hline
        Depth & 1 & 2 & 3 & 4 & 5 \\
        \hline
        IoU (\%) & 59.7 & 60.1 & 60.0 & \textbf{60.3} & 60.0 \\ 
        mIoU (\%) & 30.1 & 29.9 & 30.3 & \textbf{30.4} & 30.3 \\
        \hline
        \end{tabular}
        }
    \label{tab:depth_sparsevfe}
\end{table}

\begin{table}[t]
    \centering
    \caption{Effect of the focal loss weight $\lambda_{focal}$ on SemanticKITTI validation set. The \textbf{bold} numbers denote the best results.}
    \scalebox{1.2}{
        \begin{tabular}{c|ccccc}
        \hline
        $\lambda_{focal}$ & 0.0 & 0.5 & 1.0 & 1.5 & 2.0 \\
        \hline
        IoU (\%) & 59.1 & 60.2 & \textbf{60.3} & 60.1 & 60.0 \\
        mIoU (\%) & 30.0 & 30.2 & \textbf{30.4} & 30.3 & 30.1 \\
        \hline
        \end{tabular}
        }
    \label{tab:eff_focal_loss}
\end{table}

\begin{table}[t]
    \centering
    \caption{Effect of the dice loss weight $\lambda_{dice}$ on SemanticKITTI validation set. The \textbf{bold} numbers denote the best results.}
    \scalebox{1.2}{
        \begin{tabular}{c|ccccc}
        \hline
        $\lambda_{dice}$ & 0.0 & 0.5 & 1.0 & 1.5 & 2.0\\
        \hline
        IoU (\%) & 59.9 & \textbf{60.4} & 60.3 & 60.3 & 59.9 \\
        mIoU (\%) & 29.9 & 30.2 & \textbf{30.4} & 30.3 & 30.3 \\
        \hline
        \end{tabular}
        }
    \label{tab:eff_dice_loss}
\end{table}

\noindent\textbf{Effect of using different resolutions of voxel-view representation.}
We investigate the effect of using different resolutions of voxel-view representation on the SemanticKITTI validation set. The voxel-view encoder employs a multi-scale design with three hierarchical voxel grids of resolutions $32\times32\times4$, $64\times64\times8$, and $128\times128\times16$. As shown in Table~\ref{tab:voxel_sizes}, this multi-scale strategy effectively captures geometric structures at different levels of detail, thereby enriching spatial awareness and enhancing geometric perception. Compared to using a single-scale voxel representation, the multi-scale approach improves the IoU of geometric occupancy by approximately 2.8\%, which leads to accurate semantic occupancy prediction.

\noindent\textbf{Effect of using different stages of range-view representation.}
We investigate the effect of using different stages of range-view representation on the SemanticKITTI validation set. The range-view encoder employs cascaded ResNet blocks to produce multi-stage feature maps, which are fused via channel-wise concatenation. As shown in Table~\ref{tab:range_layers}, deeper feature consistently yield higher performance owing to their richer semantic representations and larger receptive fields. However, shallow features retain fine-grained spatial details that are crucial for precise scene reconstruction. By fusing shallow and deep features, the multi-stage fusion effectively combines high-resolution spatial details from early layers with high-level semantic context from deeper layers. Specifically, the multi-stage fusion outperforms other single-stage variants by approximately 1.8\% in semantic mIoU.

\noindent\textbf{Effect of the number of voxel-to-range fusion modules.}
We study the impact of the number of voxel-to-range fusion modules on the SemanticKITTI validation set. As shown in Table~\ref{tab:num_deformattn}, increasing the number of deformable attention layers progressively enhances the interaction between voxel- and range-view features, leading to improved semantic understanding. We apply three modules for feature fusion while balancing model performance and computational burden.

\noindent\textbf{Effect of the depth of range-to-voxel fusion modules.}
We study the impact of the depth of the range-to-voxel fusion modules on the SemanticKITTI validation set. As shown in Table~\ref{tab:depth_sparsevfe}, stacking more sparse basic blocks progressively improves the model's geometric perception through richer voxel-level feature extraction. We set the depth of the range-to-voxel fusion module to four, which achieves a favorable balance between performance and efficiency.

\begin{figure}[h]
    \centering
    \includegraphics[width= 1.0\linewidth]{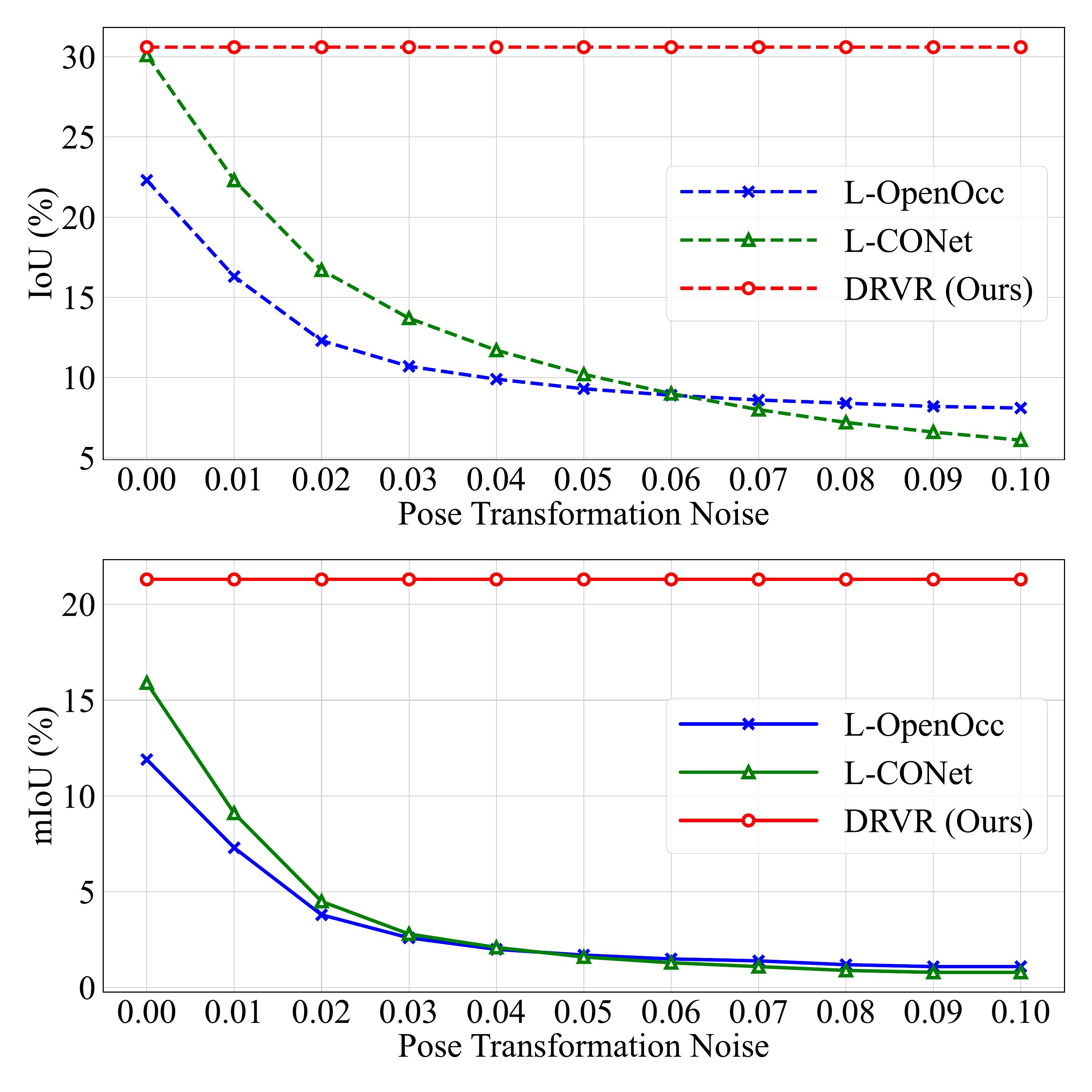}
    \vspace{-0.6cm}
    \caption{Comparisons of the model performance under different levels of pose transformation noise on nuScenes-Occupancy validation set.}
    \label{fig:pose_noise_analysis}
\end{figure}

\begin{figure}[h]
    \centering
    \includegraphics[width= 1.0\linewidth]{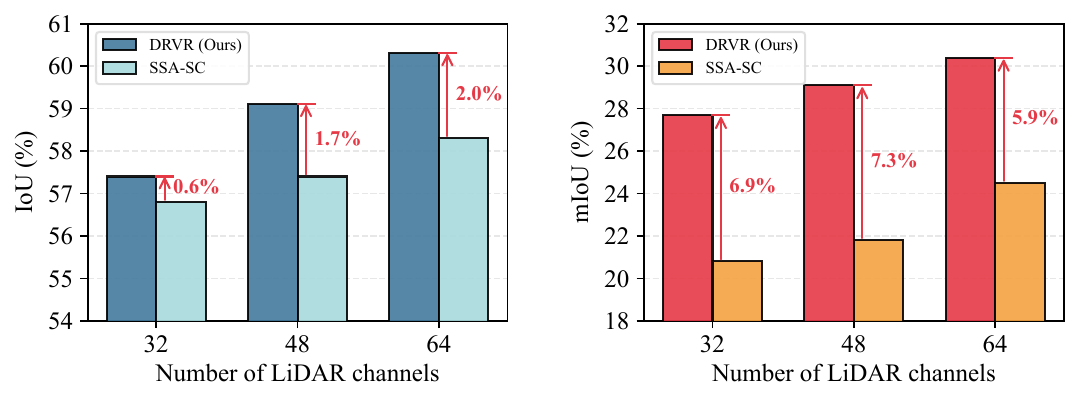}
    \vspace{-0.6cm}
    \caption{Comparisons of the model performance under different numbers of LiDAR channels on SemanticKITTI validation set.}
    \label{fig:lidar_channels_analysis}
\end{figure}

\noindent\textbf{Effect of hyperparameters in the objective function.}
We investigate the effect of hyperparameters in the objective function on the SemanticKITTI validation set. Specifically, we fix $\lambda_{dice}$ to 1.0 and vary the focal loss weight $\lambda_{focal}\in \{0.0, 0.5, 1.0, 1.5, 2.0\}$. As shown in Table~\ref{tab:eff_focal_loss}, incorporating focal loss significantly enhances the model's geometric perception capability and achieves peak performance at $\lambda_{focal} = 1.0$. Moreover, we evaluate the effect of dice loss by setting $\lambda_{focal}$ to 1.0 and varying $\lambda_{dice}\in \{0.0, 0.5, 1.0, 1.5, 2.0\}$. As shown in Table~\ref{tab:eff_dice_loss}, introducing dice loss consistently improves both geometry IoU and semantic mIoU, increasing from 59.9\% and 29.9\% (without dice loss) to 60.3\% and 30.4\% at $\lambda_{dice} = 1.0$.

\subsection{Robustness against Pose Transformation Noise}
\label{robustness_to_noise}
On nuScenes-Occupancy benchmark, given the challenge of inferring the coherent semantics and geometry of 3D space from sparse point clouds, existing methods~\cite{wang2023openoccupancy, wang2024occgen} stack multi-sweep point clouds for dense spatial information. However, these methods rely on the corresponding pose transformation matrix to achieve accurate alignment between the multi-sweeps and may be affected by the pose transformation noise in practice. To verify this, we apply different levels of Gaussian noise to the pose transformation matrix. Specifically, given the pose transformation matrix $\bR \in \mathbb R^{4 \times 4}$, we adopt the noise by $\bR_{noise} = \bR + e$, where $e\sim N(\mu, \sigma^2)$, $\mu=0$ and $\sigma\in \{0.00, 0.01, 0.02, \ldots, 0.10 \}$. As shown in Figure~\ref{fig:pose_noise_analysis}, as multi-sweep methods heavily rely on the pose transformation matrix, their model performance is degraded with the increase of Gaussian noise. In contrast, single-sweep methods (\textit{e.g.}, our \shortname) avoid the need for multi-sweep point cloud alignment and demonstrate stronger robustness to perturbations in the pose transformation matrix.

\subsection{Effectiveness under LiDAR Sparsity}
SemanticKITTI is captured using a 64-line LiDAR, providing sufficiently dense point clouds that enable existing methods to process single-sweep inputs directly. While our method already achieves state-of-the-art performance on such dense data, we further investigate its effectiveness in sparser scenarios. To this end, we simulate 32-line and 48-line LiDAR scans by downsampling the original point clouds along the vertical direction, and train both our \shortname and SSA-SC~\cite{yang2021semantic} using these sparse point clouds as input. As shown in Figure~\ref{fig:lidar_channels_analysis}, our \shortname consistently outperforms SSA-SC across different number of LiDAR channels, demonstrating stable performance under varying levels of point cloud sparsity. Specifically, our \shortname surpasses SSA-SC by 0.6\%, 1.7\%, and 2.0\% in geometry IoU and 6.9\%, 7.3\%, and 5.9\% in semantic mIoU under 32-, 48-, and 64-line settings, respectively. The IoU gap narrows with decreasing the number of LiDAR channels, constrained by the inherent scarcity of geometric information in extremely sparse point clouds. In contrast, our \shortname with only 32 lines achieves a higher mIoU than SSA-SC using the full 64-line input, indicating its ability to maintain superior semantic understanding in highly sparse sensing conditions. Furthermore, the larger mIoU gap observed under sparser settings (6.9\%, 7.3\%) compared to the 64-line case (5.9\%) highlights the effectiveness of our dual range-voxel representation in mitigating semantic context degradation. These results underscore the strength of our range-voxel fusion design in jointly enhancing geometric occupancy and semantic prediction under real-world resource-constrained sensing scenarios.

\section{Conclusion}
\label{sec:conclusion}
In this work, we propose a dual range-voxel representation that leverages the range-view context and voxel-view geometry of single-sweep sparse and incomplete point clouds for 3D semantic occupancy prediction, eliminating the efficiency and robustness concerns associated with the multi-sweeps. Specifically, to alleviate the spatial discontinuity and contextual sparsity of point clouds, we perform spherical projection to obtain the compact and continuous range-view image for semantic representation learning. To fully exploit the spatial information of point clouds, we design the geometry-aware voxel-view encoder to extract multi-scale features and merge them for geometric occupancy prediction. Moreover, we propose the range-voxel fusion module to effectively cooperate range- and voxel-view features via voxel-to-range and range-to-voxel fusions. The experimental results on three benchmarks show the superiority of our method. Note that fusing complementary information from multi-modal sensors effectively improves model robustness against bad weather. In the future, we will extend our method to use multi-modal data and enhance its robustness under varying weather conditions.

\ifCLASSOPTIONcaptionsoff
  \newpage
\fi
\bibliographystyle{IEEEtran}
{
 \bibliography{arxiv}
}

\end{document}